# Localization of Pallets on Shelves Using Horizontal Plane Projection of a 360-degree Image

Yasuyo Kita, Yudai Fujieda, Ichiro Matsuda and Nobuyuki Kita

*Abstract*— In this paper, we propose a method for calculating the three-dimensional (3D) position and orientation of a pallet placed on a shelf on the side of a forklift truck using a 360-degree camera. By using a 360-degree camera mounted on the forklift truck, it is possible to observe both the pallet at the side of the forklift and one several meters ahead. However, the pallet on the obtained image is observed with different distortion depending on its 3D position, so that it is difficult to extract the pallet from the image. To solve this problem, a method [1] has been proposed for detecting a pallet by projecting a 360-degree image on a vertical plane that coincides with the front of the shelf to calculate an image similar to the image seen from the front of the shelf. At the same time as the detection, the approximate position and orientation of the detected pallet can be obtained, but the accuracy is not sufficient for automatic control of the forklift truck. In this paper, we propose a method for accurately detecting the yaw angle, which is the angle of the front surface of the pallet in the horizontal plane, by projecting the 360-degree image on a horizontal plane including the boundary line of the front surface of the detected pallet. The position of the pallet is also determined by moving the vertical plane having the detected yaw angle back and forth, and finding the position at which the degree of coincidence between the projection image on the vertical plane and the actual size of the front surface of the pallet is maximized. Experiments using real images taken in a laboratory and an actual warehouse have confirmed that the proposed method can calculate the position and orientation of a pallet within a reasonable calculation time and with the accuracy necessary for inserting the fork into the hole in the front of the pallet.

## I. INTRODUCTION

In the logistics field, where demand continues to increase, automation of work is strongly desired due to labor shortages. Among various problems existing, automatic operation of forklifts in warehouses is also a major problem. In forklift operations, a load is placed on a pallet, and a fork is inserted into a hole on the side of the pallet to lift and move the load. Therefore, in order to automate forklift operations, technology for automatically detecting the three-dimensional (3D) position and orientation of the target pallet is indispensable.

For this reason, many studies on automatic position and orientation detection of pallets by visual recognition have been conducted in recent years. Although some studies have made special fiducial marks on pallets to facilitate visual recognition [2, 3], marking a large number of pallets is a time-consuming task, and there is a problem that the markers are damaged by

Y. Kita and N. Kita are with TICO-AIST Cooperative Research Laboratory for Advanced Logistics, National Institute of Advanced Industrial Science and Technology (AIST), Tsukuba, Japan (e-mail: y.kita@aist.go.jp).
Y. Fujieda and I. Matsuda are with Faculty of Science and Technology, Tokyo University of Science, Noda, Chiba, Japan.

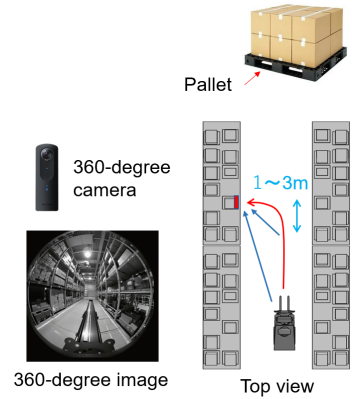

Figure 1. Purpose of the present paper: localization of a pallet of interest so that the forklift properly turns the wheel to face the pallet directly.

use and become unusable. For this reason, more studies have been made on methods without markers; two-dimensional (2D) images [4, 5], 2D LIDAR [6, 7], 3D sensors [8, 9], and combinations thereof have been used as inputs.

However, many of these studies assume a situation in which a forklift approaches a pallet placed on the floor from a direction close to the front. On the other hand, in the case of unloading a pallet placed on a shelf by a forklift moving in a passage between shelves, as shown in Fig. 1, the forklift must proceed along the shelf from the side at a 90-degree angle from the front direction of the pallet, turn the handle 1 to 3m before the pallet, and face toward the pallet so that the fork enters the hole of the pallet. To do this, it is necessary to grasp the 3D position and orientation of the pallet a few meters before the pallet. Since a wide field of view is necessary to observe both the pallet next to the forklift and the pallet a few meters ahead, we proposed to mount a 360-degree camera as shown in Fig. 1 on the forklift [1]. However, in a 360-degree image, the pallet is distorted differently depending on the position, so that it is difficult to detect the pallet and its 3D position and orientation directly from this image. Therefore, taking advantage of the fact that the front surface of the pallet and the shelf are close to the same plane, we proposed a method to project a 360-degree image on the vertical plane that is the front surface of the shelf and to detect the pallet from the projected image using a full-scale pallet front model. Further, the angle and position of the vertical projection plane are changed in the vicinity thereof. The 3D position and orientation of the pallet are detected by finding the projection plane that best matches the model [10]. However, it is a practical problem that it takes a long time to calculate the position and orientation, and there are occasional errors in the calculation of the yaw angle, which is the rotation around the vertical axis.

This paper proposes a method for calculating the yaw angle of a pallet by projecting a 360-degree image on a horizontal plane such as the bottom surface of the pallet to generate an image of the boundary of the pallet existing on the horizontal plane as viewed from directly above. Then, the position of the pallet is also determined by finding the position of the vertical plane of the calculated yaw angle where the projected image of the front surface of the pallet coincides most with the full-size model of the front surface of the pallet. By this method, we aim to make the 3D localization process of the pallet using the 360-degree image robust and fast.

## II. Related Works

Byun et al. [4] proposed a method for calculating the yaw angle of a pallet by detecting the pallet from a 2D image and then detecting the inclination of a plane in which the upper and lower lines of the pallet are parallel while changing the inclination of a virtual vertical projection plane. However, it is assumed that the pallet can be separated from the background due to the difference in color, and that the pallet faces almost directly in front of the camera. Varga et al.[5] devised image features, such as "normalized pair differences," that were invariant to certain types of illumination changes. Baglivo et al.[6] used both a 2D LRF (Laser Range Finder) and a color camera to detect pallets. If three collinear straight segments were found by a LRF scan, an initial guess about the relevant pallet's position was generated and verified using color images.

In recent years, many studies have used deep learning for palette detection. Mohamed et al. [7] proposed a two-step method in which the 2D LRF data is input as a 2D image of a room viewed from above, palette candidate regions are detected using Faster R-CNN, and the candidates are narrowed down by CNN. However, the reliability is still uncertain, and the results of input images in time series must be integrated by a Kalman filter-based method to produce the final result. Moreover, the accuracy of position detection is low, and 2D LRF can only handle objects of fixed height, such as objects placed on the floor. Ito et al. [8] developed a sensor that can obtain range images and grayscale images, and applied DCNN. In doing so, the accuracy was improved by treating the peak intensity image as the certainty of the range data. Research was also conducted using an inexpensive RGB-D camera as an input. Li et al. [9] compared several deep learning methods and applied SSD. However, when an RGB-D camera is used, the pallet is limited to a place that is substantially in front and not too far away, which is the measurement range of the camera.

All these methods are effective only when vehicles with sensors are close and approaching from the front direction of a target pallet. As described in Section 1, our goal has different situation from these methods. When a forklift approaches along the shelf from a direction at 90 degrees from the front direction of the pallet, the direction of the line of sight to the target pallet changes greatly as the forklift moves, as shown by the blue arrows in Fig. 1. Therefore, the camera mounted on the forklift is required to have a wide field of view. When a 360-degree camera as shown in Fig. 1 is mounted on a forklift, this condition is satisfied. However, on a 360-degree image, the pallet is observed with different distortion depending on the position, so that it is difficult to detect the pallet directly from the image.

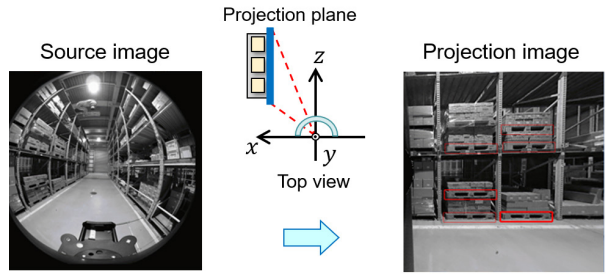

Figure 2. Projection to the front surface of the shelf.

To solve this problem, we proposed a method in which a 360-degree image is projected on the vertical plane of the front surface of the shelf to detect the pallet by template matching using a full-scale pallet front model from the projected image, taking advantage of the fact that the front surface of the shelf is close to the front surface of the pallet in terms of orientation and position [1]. At the same time as the detection, the 3D position and orientation of the pallet can be calculated from the three-dimensional equation of the projection plane and the 2D position on the projection image. However, this is an approximate value because the projection plane does not completely coincide with the front surface of the pallet. After that, the final localization is performed by finding the position and orientation where the projected image is most consistent with the full-scale model of the front surface of the pallet while changing the orientation and position of the projected vertical plane in the neighborhood [10]. However, since the change in the projection image due to the difference in the yaw angle is small, there is a problem that it is difficult to robustly extract the projection plane having the maximum matching degree. Furthermore, as shown in [10], the change in the projected image caused by the difference in the yaw angle and that caused by the difference in the distance from the camera are confusing. Therefore, the degree of coincidence with the pallet front surface model becomes a gentle peak in the shape of a ridge with respect to these changes. As a result, the combination of the angle and the distance that most coincides does not become a clear point. This causes an error in the result, and there is a problem that it takes a lot of time to correctly detect the peak in such a situation. Therefore, we propose a method to explicitly determine the yaw angle by utilizing the projection onto the horizontal plane where the yaw angle difference appears more clearly, and to determine the position under the condition of the obtained yaw angle. This method improves the accuracy of the position and orientation calculation and also reduces the calculation time.

## III. Method

### A. Pallet Detection

For pallet detection, the method of [1] is used. Specifically, the equation of the vertical plane of the shelf's front surface (blue line in the top view of Fig. 2) is first obtained. Then, a 360-degree image is projected on the vertical plane to obtain an image as shown in the right view of Fig. 2. Details of how to calculate the intensity (color) of the pixel on the projection image from the source image are described in [1]. Since we can assume with some allowance that the front surfaces of the pallets are coplanar to the front surface of the shelves, the pallet front plane is observed to have almost the same size and shape as the actual pallet on this plane. Therefore, the pallet can be detected by template matching using the model of the pallet front plane as a template. The detection result example is shown by the red frames in the right figure of Fig. 2. The approximate position and orientation of the pallet can be calculated from the detected position on the projected image plane.

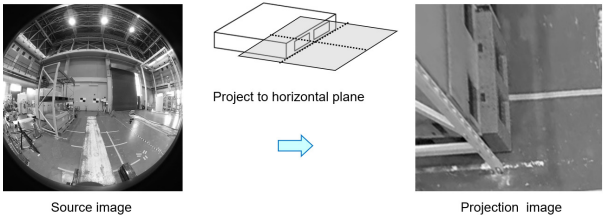

Figure 3. Projection to the horizontal plane of the pallet bottom.

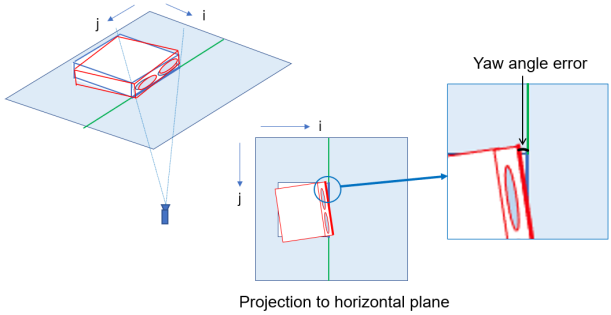

Figure 4. Principle of detecting yaw angle error.

*B. Calculation of Yaw Angle*

Based on the initial position and orientation of the pallet, a 360-degree image is projected onto the horizontal plane of the bottom surface (or the top surface) of the pallet as shown in Fig. 3. At this time, the projection plane is set so that the vertical center line of the image (the green line in Fig. 4) coincides with the intersection line between the front surface and the bottom surface of the pallet (the blue thick line in Fig. 4). By projecting in this way, the same effect as projecting from directly above can be obtained for the line existing on this projection horizontal plane. If the initial orientation (the blue line in Fig. 4) is correct, the intersection line on the projected image coincides with the center line. However, if the orientation is actually deviated as shown by the red line in Fig. 4, the intersection line and the center line form an angle corresponding to the deviation. Therefore, the correct angle can be calculated by correcting the yaw angle by this angle.

The height of the horizontal projection plane is determined depending on the position of the pallet relative to the camera. If the pallet is positioned below the camera, the lower boundary of the pallet is in front of the adjacent objects and represents the orientation of the pallet. As a result, the horizontal plane should be set to the bottom plane of the pallet which includes this boundary line. Conversely, if the pallet is higher than the camera, the top surface of the pallet should be used. When the boundary line with the background is difficult to detect, a horizontal plane including the boundary line of the hole on the front surface of the pallet may be used. If the camera and the pallet are at approximately the same height, the horizontal plane image cannot be clearly obtained, and this method cannot be applied.

*C. Calculation of Position*

The vertical projection plane with the corrected yaw angle is moved forward and backward with respect to the camera to determine the position that gives the projection image most consistent with the pallet front model. For the calculation of the degree of matching, the edge template considering the round corners of the pallet proposed in [10] is used.

IV. EXPERIMENTS

The performance of the proposed method is investigated using real images taken in a laboratory and a real warehouse.

*A. Experiments using Laboratory Data*

As shown in Fig. 5 (a), a shelf on which pallets were stacked was prepared in the laboratory. The experiment was carried out using the image taken by RICOH ThetaV [11] mounted on the forklift which was approaching the target pallet by going straight on the aisle. A total of 180 images were taken while the distance along the shelf to the pallet approached from 4.7m to 1.4m. These were culled and 82 images were used as independent input images. By using the B signal (Blue signal) for a blue pallet placed on the floor, the boundary between the pallet and the background is clear as shown in Fig. 5 (b). Using this experimental data, an experiment was conducted on the accuracy of the yaw angle calculation when the boundary was correctly obtained. The results obtained by the method proposed in [1] were used as the initial position and orientation for setting the horizontal plane. The horizontal projection plane was set at a position including the bottom surface of the pallet. The boundary between the pallet and the background was extracted by performing edge detection, then performing straight line detection by Hough transform, and selecting the line closest to the center line. Fig. 5(b) shows an example of the detection results, in which red lines and the blue line are lines extracted by Hough transform and the line extracted as a boundary line respectively. Fig. 5(c) is an enlarged view thereof. In this example, a yaw angle deviation of 3.6 degrees is detected. After the yaw angle is corrected by this degree, the horizontal plane projection is performed again to make sure the effect. The resulting projected image is shown in Fig. 5 (d), in which the green center line coincides with the boundary line of the projected image.

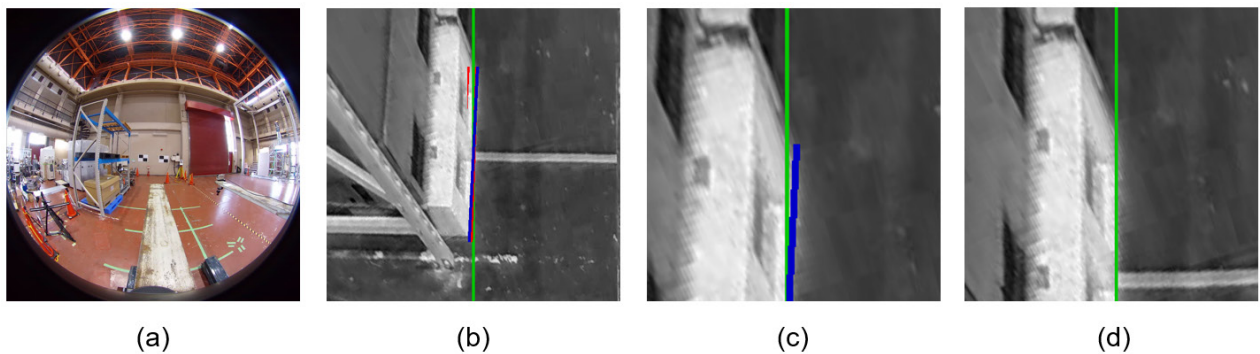

Figure 5. Example of experimental results of correcting yaw angle error: (a) source image; (b) projection image to the horizontal plane (green line is the center line of the image, blue line is detected boundary line); (c) angular deviation in the enlarged view (3.6 degree); (d) projection image to the horizontal plane after correcting angular deviation.

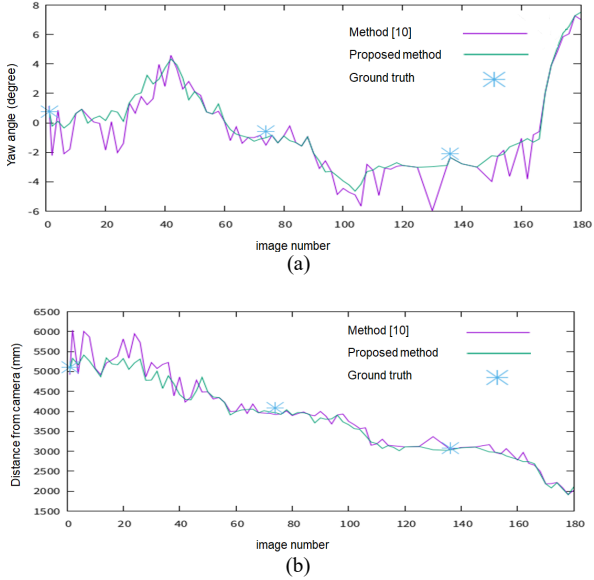

Figure 6. Experimental Result 1: (a) results of yaw angle; (b) results of distance from the camera.

The experimental results of all the images used in the experiment are shown in Fig 6. The purple line shows the results calculated by [10], and the green line shows the results corrected by the proposed method. On the way, the forklift was stopped at three places to calculate the true value of the yaw angle and position using a total station surveying instrument, Leica [12]. The results are indicated by blue asterisks in the graph. In this experiment, the forklift approached the pallet almost parallel to the front surface of the shelf, and the yaw angle with respect to the camera coordinate system changed smoothly around zero degree.

However, as shown in the graph of the yaw angle in Fig. 6 (a), in the conventional method, the calculation result is jaggy and unstable. In actual application, accuracy within ± two degrees was required, but there were fluctuations of three degrees in some places. On the other hand, as a result of correcting the yaw angle by using the proposed method, the graph becomes smooth like the green line, which almost coincides with the true values. The fluctuation is almost within ± one degree.

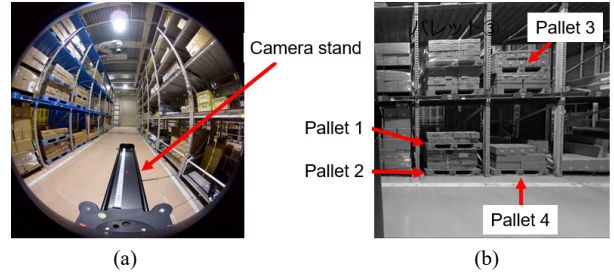

Figure 7. Experimental condition:(a) Source image;(b) pallets used in the experiment.

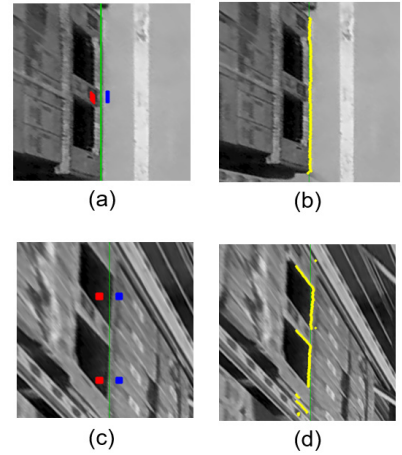

Figure 8. Process of extracting a boundary line: (a), (c) regions for calculating threshold; (b), (d) candidates of points on the boundary line.

In Fig. 6 (b), the green line indicates the position of the pallet determined from the position of the vertical plane most coincident with the pallet front model using the corrected yaw angle. In the existing method in which the position and the yaw angle are determined at the same time based on the coincidence degree, there is a lot of fluctuation (purple line). However, in the modification by the proposed method, the result is smoother (green line). In particular, a large fluctuation is not observed in the data number 60 or more where the distance to the pallet is closer than 4m.

*B. Experiments using Warehouse Data*

Using the data of the actual warehouse image, the experiment including whether the boundary line in the horizontal projection image can be stably detected was carried out. While moving the camera at 100 mm intervals along the shelf using a camera stand and slider with a scale, wide-angle views such as those shown in Fig. 7(a) were recorded. We focused on four pallets on the left side shown in Fig.7(b). For pallets 1, 2, and 4 located below the camera, a horizontal plane was set on the lower surface of the pallet. For pallet 3 located above the camera, the projection was made on a horizontal plane including the upper edge of the pallet front hole where a clear edge appeared.

In the boundary line extraction process, as shown in Figs. 8 (a) and 8 (c), the average density value of the two small regions (the red region and the blue region in the figures) sandwiching the center line is calculated. By using the intermediate value as a threshold value for raster scanning, boundary point candidates in each row of the image are extracted (the yellow points in Figs. 8 (b) and 8 (d)). Then, a straight line is detected by Hough transform. Fig. 9 shows an example in which the yaw angle is corrected by using the upper edge of the hole of the pallet 3 as the boundary line. In the initial position and orientation (Fig. 9 (a)), the angular deviation between the center line (the green line) and the upper line of the hole (the blue line) is detected by 2.8 degrees (Fig. 9 (b)). Fig. 9 (c) shows a horizontal plane projection image created on the basis of the result after correcting the yaw angle by this angle. Regarding the pallets 2 and 4, the results of the yaw angle were obtained stably even by the existing method; the accuracy of the yaw angle did not change much. However, regarding the pallets 1 and 3, there was a case in which the yaw angle deviated by nearly 3 degrees as shown by the purple line in the graph of Fig. 10. By correcting the yaw angle by the horizontal plane projection, the results were obtained stably as shown by the green line.

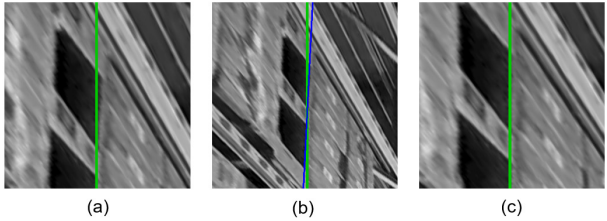

Figure 9. (a) horizontal projection image at initial state; (b) detected boundary line (blue line, 2.8 degree deviation); (c) horizontal projection image after correcting the deviation.

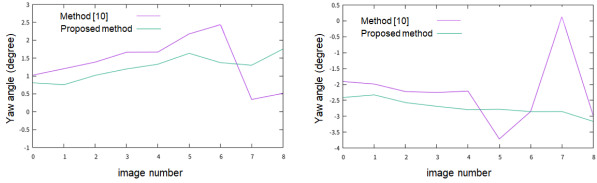

Figure 10. Results of yaw angle: (a) results of Pallet 1; (b) results of Pallet 3.

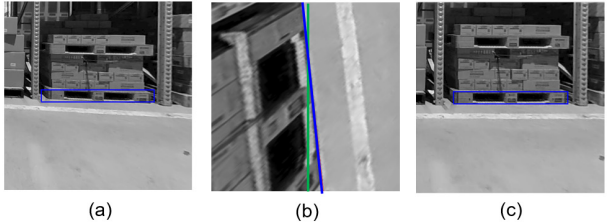

Figure 11. Correction from largely deviated initial state: (a) projection to the front surface of the pallet at initial state; (b) projection to the horizontal plane; (c) projection to the front surface of the pallet after correcting orientation and position.

An experiment was also conducted to check the allowable range of deviation of the initial position and orientation from the true value of the pallet front surface. It was confirmed that the proposed method can correct the position if the initial projected plane is within the yaw angle error of ± 5 degrees and the position error of -10cm to + 50 cm from the true value. This allowable error is sufficiently larger than the expected difference between the front surface of the shelf and the front surface of the pallet. Fig. 11 shows an example in which the initial position and orientation are intentionally greatly shifted from the true value; while the true position is (2027, -1521, -760) (mm), the initial position was shifted by about 30 cm and (1718, -1280, -642) (mm) and the initial orientation was set 5 degrees. The image projected to the front surface of the pallet based on this initial value is as shown in Fig. 11 (a), which is largely different from the actual pallet model shown by the blue line. The horizontal projected image calculated based on the initial position and orientation was as shown in Fig. 11 (b). The yaw angle was corrected to -1.5 degree from the inclination of the detected blue boundary line. The distance to the vertical plane with a yaw angle of -1.5 degree was changed, and the position of the pallet was calculated to be (2033, -1521, -760) (mm) from the position where the projection image coincides most with the front model of the pallet as shown in Fig. 11 (c). The result shows that the difference from the true position value is within the target of about 2 cm.

The calculation time for calculating the final position and orientation from the initial position and orientation was compared between the conventional method [10] and the proposed method on an Intel Core i7 2.6-GHz machine. In the conventional method, it took 16 to 27 seconds to search for a combination that maximizes the degree of coincidence while simultaneously changing the yaw angle and position. In contrast, the proposed method can calculate the results in a constant processing time of about 1.6 seconds, which is 0.16 seconds for the calculation of yaw angle and 1.41 seconds for the calculation of position. The horizontal plane projection not only improved the accuracy but also greatly improved the processing time to be acceptable for practical applications.

## V. CONCLUSION

We proposed a method to directly calculate the yaw angle of a pallet by projecting 360-degree image onto a horizontal plane including a boundary line on the front surface of the pallet. By moving the vertical plane of the calculated yaw angle back and forth with respect to the camera to obtain the position that gives the projected image that most matches the actual size of the front of the pallet, the position of the pallet can also be determined in a short time. Experiments using real images confirmed that the initial position and orientation approximated from the front surface of the shelf can be used to set the initial horizontal plane. It was also confirmed that the accuracy of the obtained results and the required calculation time are within the allowable range when considering actual application. Future issues include the development of a method for stably extracting a boundary line from the projected image when calculating the yaw angle under various real environments.